\documentclass{article}
\PassOptionsToPackage{numbers}{natbib}
 \usepackage[final]{nips_2017}

\usepackage[utf8]{inputenc} 
\usepackage[T1]{fontenc}    
\usepackage{hyperref}       
\usepackage{url}            
\usepackage{booktabs}       
\usepackage{amsfonts}       
\usepackage{nicefrac}       
\usepackage{microtype}      
\usepackage{macros}
\usepackage{tabularx}
\usepackage{array}
\usepackage{amsmath, amsfonts, amssymb}

\title{Using Noisy Extractions to \\ Discover Causal Knowledge}

\author{
	Dhanya Sridhar \\
	University of California Santa Cruz\\
	\texttt{dsridhar@soe.ucsc.edu} \\
	 \And
	 Jay Pujara \\
	 Information Sciences Institute \\
	 \texttt{jay@cs.umd.edu} \\
	 \And
	 Lise Getoor \\
	 University of California Santa Cruz\\
	 \texttt{getoor@soe.ucsc.edu} \\
}

\begin{document}
	
	\maketitle

\section{Introduction}
\label{sec:intro}

Knowledge bases (KB) constructed through information extraction from text play an important role in query answering and reasoning. 
Since automatic extraction methods yield results of varying quality, typically, only high-confidence extractions are retained, which ensures precision at the expense of recall. 
Consequently, the noisy extractions are prone to propagating false negatives when used for further reasoning. 
However, in many problems, empirical observations of entities, or observational data, are readily available, potentially recovering information when fused with noisy extractions. For reasoning tasks where both empirical observations and extractions can be obtained, an open and critical problem is designing methods that exploit both modes of identification.  

In this work, we study a particular reasoning task, the problem of discovering causal relationships between entities, known as causal discovery. 
There are two contrasting types of approaches to discovering causal knowledge. One approach attempts to identify causal relationships from text using automatic extraction techniques, while the other approach infers causation from observational data.
For example, prior extraction-based approaches have mined causal links such as regulatory relationships among genes directly from scientific text \cite{poon2014literome,song2009text}.
However, the extracted links often miss complex and longer-range patterns that require observational data.
On the other hand, given observations alone, extensive work has studied the problem of inferring a network of cause-and-effect relationships among variables \cite{spirtes1991algorithm,chickering1996learning}. Observational data such as gene expression measurements are used to infer causal relationships such as gene regulation \cite{magliacane2016ancestral}.
Prior approaches use constraints to find valid causal orientations from observational data \cite{hyttinen2013discovering,hyttinen2014constraint,magliacane2016ancestral,spirtes1991algorithm,claassen2012logical}.
Although the constraints offer attractive soundness guarantees, the need for observed measurements of variables remains costly and prohibitive when experimental data is unpublished.
Extractions such as interactions between genes mined directly from text provide a coarse approximation of unseen observational data.
Combining extractions mined from KBs with observed measurements where available to for \cd~can alleviate the cost of obtaining experiment-based data.

We propose an approach for fusing noisy extractions with observational data to discover causal knowledge.
We introduce \ourmethod, a probabilistic model over causal relationships that combines commonly used constraints over observational data with extractions obtained from a KB.
\ourmethod~uses the probabilistic soft logic (PSL) modeling framework to express causal constraints in a natural logical syntax that flexibly incorporates both observational and KB modes of evidence. 
As our main contributions:
\begin{enumerate}
	\item We introduce the novel problem of combining noisy extractions from a KB with observational data.
	\item We propose a principled approach that uses well-studied \cd~constraints to recover long-range patterns and consistent predictions, while cheaply acquired extractions provide a proxy for unseen observations.
	\item We apply our method gene regulatory networks and show the promise of exploiting KB signals in causal discovery, suggesting a critical, new area of research.
\end{enumerate}

We compare \ourmethod~with a conventional logic-based approach that uses only observational data to perform \cd. We evaluate both methods on transcriptional regulatory networks of yeast. 
Our results validate two strengths of our approach: 1) \ourmethod~achieves comparable performance with the well-studied conventional method, suggesting that noisy extractions are useful approximations for unseen empirical evidence; and 2) global logical constraints over observational data enforce consistency across predictions and bolster \ourmethod~to perform on par with the competing method.
The results suggest promising new directions for integrating knowledge bases in causal reasoning, potentially mitigating the need for expensive observational data.
\section{Background on Logical Causal Discovery}
\label{sec:background}


The inputs to traditional causal discovery methods are $m$ independent observations of $n$ variables $\mathbf{V}$. The problem of causal discovery is to infer a directed acyclic graph (DAG) $\mathcal{G^*} = (\mathbf{V}, \mathbf{E})$ such that each edge $E_{ij} \in \mathbf{E}$ corresponds to $V_i$ being a \emph{direct cause} of $V_j$, where changing the value of $V_i$ always changes the value of $V_j$. 

Since graphical model $\mathcal{G^*}$ encodes conditional independences among $\mathbf{V}$, \cd~algorithms exploit the mapping between observed independences in the data and paths in $\mathcal{G^*}$ to specify constraints on the output.
The PC algorithm \cite{spirtes1991algorithm} is a canonical such method that performs independence tests on the observations to rule out invalid causal edges. Constraints over causal graph structure can also be encoded with logic \cite{hyttinen2013discovering,hyttinen2014constraint,magliacane2016ancestral}. In a logical \cd~system, independence relations are represented as logical atoms. Logical atoms consist of a predicate symbol $p_i(\cdot)$ with $i$ variable or constant arguments and take boolean or continuous truth values. To avoid confusion with logical variables, for the remainder of this paper, we refer to $V \in \mathbf{V}$ as vertices. As inputs to logical \cd, we require the following predicates to represent the outcomes of independence tests among $\mathbf{V}$:
\begin{enumerate}
	\item [--]$\textsc{Dep}(A, B)$, $\textsc{Indep}(A, B)$ refers to statistical (in)dependence between vertices $V_A$ and $V_B$ as measured by the independence test $V_A \indep V_B$. The conditioning set is the empty set.
	\item[--]$\textsc{CondDep}(A, B, S)$, $\textsc{CondIndep}(A, B, S)$ corresponds to statistical (in)dependence between vertices $V_A$ and $V_B$ when conditioned on set $\mathbf{S} \subset \mathbf{V}\setminus\{V_A, V_B\}$. The independence test $V_A \indep V_B | \mathbf{S}$ is performed.
\end{enumerate}

The outputs are of a logical \cd~system are represented by the following \emph{target} predicates:
\begin{enumerate}
	\item [--]$\textsc{Causes}(A, B)$ refers to the absence or presence of a causal edge between $V_A$ and $V_B$, and is substituted with all pairs of vertices $A, B \in \mathbf{V}$. Finding truth value assignments to these atoms is the goal of causal discovery.
	\item [--]$\textsc{Ancestor}(A, B)$ corresponds to the absence or presence of an \emph{ancestral} edge between all vertices $V_A$ and $V_B$, where $V_A$ is an ancestor of $V_B$ if there is a directed causal path from $V_A$ to $V_B$. We may additionally infer the truth values of ancestral atoms jointly with causal atoms.
\end{enumerate}

Given the independence tests over $\mathbf{V}$ as input, the goal of logical \cd~is to find consistent assignments to the causal and ancestral output atoms.

\section{Using Extractions in Causal Discovery}
\label{sec:problem}

In the problem of fusing noisy extractions with \cd, in addition to the observations, we are given a set of variables $\mathbf{K} = \{K_{11} \ldots K_{nn}\}$ of evidence from knowledge base (KB) $\mathcal{K}$, where $K_{ij}$ is an affinity score of the interaction between $V_i$ and $V_j$ based on text extraction.

Extending previous logical \cd~methods, we additionally represent $\mathbf{K}$ in the predicate set with $\textsc{TextAdj}(A, B)$. 
$\textsc{TextAdj}(A,B)$ corresponds to $K_{AB}$ and denotes the absence or presence of an undirected edge, or adjacency, between $V_A$ and $V_B$ as extracted from text. 
Evidence of adjacencies is critical to inference of $\textsc{Causes}(A, B)$.
However, adjacencies in standard \cd~are inferred from statistical tests alone.
In our approach, we replace statistical adjacencies with $\textsc{TextAdj}(A,B)$.
The goal of fusing KB evidence in logical \cd~is to find maximally satisfying assignments to the unknown causal atoms based on constraints over both independence and text-based signals.
In section \ref{sec:approach}, we present a probabilistic logic approach defining constraints using statistical and KB evidence.

\section{A Probabilistic Approach to Inferring Causal Knowledge}
\label{sec:approach}

Our approach uses probabilistic soft logic (PSL) \cite{bach:jmlr17} to encode constraints for causal discovery. A key advantage of PSL is exact and efficient MAP inference for finding most probable assignments. We first review PSL and then present our novel encoding constraints that combine statistical and KB information.


\subsection{Probabilistic Soft Logic}
\label{subsec:hlmrfs}
PSL is a probabilistic programming framework where random variables are represented as logical atoms and dependencies between them are encoded via rules in first-order logic.
Logical atoms in PSL take continuous values and logical satisfaction of the rule is computed using the Lukasiewicz relaxation of Boolean logic.
This relaxation into continuous space allows MAP inference to be formulated as a convex optimization problem that can be solved efficiently.

Given continuous evidence variables $\mathbf{X}$ and unobserved variables $\mathbf{Y}$, PSL defines the following Markov network, called a hinge-loss Markov random field (HL-MRF), over continuous assignments to $\mathbf{Y}$:
\begin{equation}
P(\mathbf{Y}=\mathbf{y}|\mathbf{X}=\mathbf{x}) = \frac{1}{\mathcal{Z}} \exp \Big (- \sum_{r=1}^M w_r \phi_r(\mathbf{y}, \mathbf{x}) \Big ) \mbox{ ,} \label{eqn:hl-mrf}
\end{equation}
where $\mathcal{Z}$ is a normalization constant, and 
$\phi_r(\mathbf{y}, \mathbf{x}) = \left( \max\{l_r(\mathbf{y}, \mathbf{x}),0 \}\right)$
is an efficient-to-optimize \emph{hinge-loss} feature function that scores configurations of assignments to $\mathbf{X}$ and $\mathbf{Y}$ as a linear function $l_r$ of the variable assignments.

An HL-MRF is defined by PSL model $\mathcal{M} = \{(R_1, w_1) \ldots (R_m, w_m) \}$, a set of $m$ weighted disjunctions, or rules, where $w_i$ is the weight of $i$-th rule. 
Rules consist of logical atoms and are called \emph{ground rules} if only constants appear in the atoms.
To obtain the HL-MRF, we first substitute logical variables appearing in $\mathcal{M}$ with constants from observations, producing $M$ ground rules. We observe truth values $\in [0, 1]$ for a subset of the ground atoms, $\mathbf{X}$ and infer values for the remaining unobserved ground atoms, $\mathbf{Y}$. The ground rules and their corresponding weights map to $\phi_r$ and $w_r$. To derive $\phi_r(\mathbf{y}, \mathbf{x})$, the Lukasiewicz relaxation is applied to each ground rule to derive a hinge penalty function over $\mathbf{y}$ for violating the rule.
Thus, MAP inference minimizes the weighted rule penalties to find the minimally violating joint assignment for all the unobserved variables:
$$\arg \min_{\mathbf{y} \in [0,1]^{n}} \sum_{r=1}^{m} w_r \max\{l_r(\mathbf{y}, \mathbf{x}), 0\} $$
PSL uses the consensus based ADMM algorithm to perform exact MAP inference.

\subsection{\ourmethod}

\ourmethod~extends constraints introduced by the PC algorithm \cite{spirtes1991algorithm}.
Whereas PC infers adjacencies from conditional independence tests, \ourmethod~uses text-based adjacency evidence in all causal constraints.
The text-based adjacency evidence bridges domain-knowledge contained in KBs with statistical tests that propagate causal information.

Figure \ref{fig:rules} shows all the rules used in \ourmethod.
The first set of rules follow directly from the three constraints introduced by PC.
We additionally introduce joint rules to induce dependencies between ancestral and causal structures to propagate consistent predictions.
We describe below how \ourmethod~rules upgrade PC to combine KB and statistical signals for causal discovery.

\paragraph{PC-inspired Rules}
PC uses conditional (in)dependence and adjacency to rule out violating causal orientations. However, in \ourmethod, all adjacencies are directly mined from a KB.
Rule C1 discourages causal edges between vertices that are not adjacent based on evidence in text. 
Rule C2 penalizes simple cycles between two vertices. Rules C3 and C4 capture the first PC rule and orient chain $V_i-V_j-V_k$ as $V_i \rightarrow V_j \leftarrow V_k$, a v-structure, based on independence criteria.
Rule C5 orients path $V_i \rightarrow V_j-V_k$ as $V_i \rightarrow V_j \rightarrow V_k$  to avoid orienting additional v-structures. 
Rule C6 maps to the third PC rule, and if $V_i \rightarrow V_j \rightarrow V_k$ and $V_i-V_k$, orients $V_i \rightarrow V_k$ to avoid a cycle.
PC applies these rules iteratively to fix edges whereas in \ourmethod, the rules induce dependencies between causal edges to encourage parsimonious joint inferences. 


\paragraph{Joint Rules}
Joint rules encourage consistency across ancestral and causal predictions through constraints such as transitivity that follow from basic definitions.
Rule J1 encodes that causal edges are also ancestral by definition and rule J2 is the contrapositive that penalizes causal edges to non-descendants. 
Rule J3 encodes transitivity of ancestral edges, encouraging consistency across predictions. 
Rule J4 infers causal edges between probable ancestral edges that are adjacent based on textual evidence. 
Rule J5 orients chain $V_i-V_j-V_k$ as a diverging path $V_i \leftarrow V_j \rightarrow V_k$ when $V_k$ is not likely an ancestor of $V_i$.
Joint rules give preference to predicted structures that respect both ancestral and causal graphs.

In our evaluation, we investigate the implications of using a noisy extraction-based proxy for adjacency and the benefits of joint modeling. 

\begin{figure*}
	\centering
	\scriptsize
	\caption{PSL rules for combining statistical tests and KB evidence in causal discovery.}
	\label{fig:rules}
	
	\begin{tabularx}{\textwidth}{c X}
		\toprule
		\textbf{Rule Type} & \textbf{Rules} \\
		\midrule
		PC-inspired Rules & C1) $\neg \textsc{TextAdj}(A, B) \rightarrow  \neg \textsc{Causes}(A, B)$ \\
		& C2) $\textsc{Causes}(A, B) \rightarrow \neg \textsc{Causes}(B, A)$ \\
		& C3) $\textsc{TextAdj}(A, B) \wedge \textsc{TextAdj}(C, B) \wedge \neg \textsc{TextAdj}(A, C) \wedge \textsc{CondDep}(A, C, S) \wedge \textsc{InSet}(B, S) \rightarrow \textsc{Causes}(A, B)$ \\
		& C4) $\textsc{TextAdj}(A, B) \wedge \textsc{TextAdj}(C, B) \wedge \neg \textsc{TextAdj}(A, C) \wedge \textsc{CondDep}(A, C, S) \wedge \textsc{InSet}(B, S) \rightarrow \textsc{Causes}(C, B)$ \\
		& C5) $\textsc{Causes}(A, B) \wedge \textsc{Dep}(A, C) \wedge \textsc{CondIndep}(A, C, S) \wedge \textsc{InSet}(B, S) \wedge \textsc{TextAdj}(B, C) \rightarrow \textsc{Causes}(B, C)$ \\
		& C6) $\textsc{Causes}(A, B) \wedge \textsc{Causes}(B, C) \wedge \textsc{TextAdj}(A, C) \rightarrow \textsc{Causes}(A, C) $ \\
		
		Joint Rules & J1) $\textsc{Causes}(A, B) \rightarrow \textsc{Anc}(A, B)$ \\
		& J2) $\neg \textsc{Anc}(A, B) \rightarrow \neg \textsc{Causes}(A, B)$ \\
		& J3) $\textsc{Anc}(A, B) \wedge \textsc{Anc}(B, C) \rightarrow \textsc{Anc}(A,C)$ \\
		& J4) $\textsc{Anc}(A, B) \wedge \textsc{TextAdj}(A, B) \rightarrow \textsc{Causes}(A,B)$ \\
		& J5) $\textsc{TextAdj}(A, B) \wedge \textsc{TextAdj}(B, C) \wedge \textsc{Dep}(A, C) \wedge \textsc{CondIndep}(A, C, S) \wedge \textsc{InSet}(B, S) \wedge \textsc{Causes}(B, A) \wedge \neg \textsc{Anc}(C, A) \rightarrow \textsc{Causes}(B, C)$ \\
		\bottomrule
	\end{tabularx}
	
\end{figure*}
\section{Experimental Evaluation}
\label{sec:eval}

Our experiments investigate the two main claims of our approach: 
\begin{enumerate}
	\item We study whether the noisy extractions are a suitable proxy for latent adjacencies and give similar performance to a conventional logic-based approach that impute adjacency values using only observations. 
	\item We understand the role of joint ancestral and causal rules over observational data in mitigating noise from the extraction-based evidence. 
\end{enumerate}

We evaluate \ourmethod~on real-world gene regulatory networks in yeast. We compare against \ourbaseline, the PSL model variant that performs prototypical causal discovery using only observational data. \ourbaseline~replaces $\textsc{TextAdj}$ with $\textsc{StandardAdj}$, adjacencies computed from conditional independence tests.


\subsection{Data}
Our dataset for evaluation consists of a transcriptional regulatory network across 300 genes in yeast with simulated gene expression from the DREAM4 challenge \cite{marbach2010revealing,prill2010towards}. 
We snowball sample 10 smaller subnetworks of sizes 20 with low Jaccard overlap to perform cross validation. 
The data contains 210 gene expression measurements simulated from differential equation models of the system. We perform independence tests on the real-valued measurements which are known to contribute numerous spurious correlations. In addition to the gene expression data, we model domain knowledge based on undirected protein-protein interaction (PPI) edges extracted from the Yeast Genome Database: 
$$ \textsc{Anc}(A, B) \wedge \textsc{Local}_\textsc{PPI}(A, B) \rightarrow \textsc{Causes}(A, B) $$

We obtain text-based affinity scores of interaction between pairs of yeast genes from the STRING database. STRING finds mentions of gene or protein names across millions of scientific articles and computes the co-occurrence of mentions between genes. As an additional step, STRING extracts relations between genes and increases the affinity score if genes are connected by salient terms such as ``binds to'' or ``phosphorylates.'' 

\subsection{Results}

\begin{table*}[th!b]
		\centering
		\begin{tabular}{l l}
		\toprule
		\textbf{Model Variant} &  \textbf{$F_{1}$}\\
		\midrule
		\ourmethod & 0.19 $\pm$ 0.08 \\
		\ourbaseline & 0.20 $\pm$ 0.05 \\
		\hline \\
		\pslpc & 0.17 $\pm$ 0.07 \\
		\baselinepc & 0.19 $\pm$ 0.05 \\
		\hline
		\bottomrule
		\end{tabular}
		\caption{\ourmethod~achieves comparable performance with \ourbaseline, suggesting that noisy extractions can approximate unseen adjacencies. Without joint rules, \pslpc~shows worse performance, pointing to the benefit of sophisticated joint modeling in mitigating noisy extractions.}
		\label{tab:results}
\end{table*}
\begin{table*}
	\centering
	\begin{tabular}{l l l}
		\toprule
		\textbf{Adjacency} &  \textbf{Precision} & \textbf{Recall}\\
		\midrule
		\textsc{TextAdj} & 0.32  & 0.11 \\
		\textsc{StandardAdj} & 0.27 & 0.3 \\
		\bottomrule
	\end{tabular}
\caption{Extraction-based adjacencies achieve higher-precision but lower recall, further substantiating need for joint rules in recovering missing causal orientations.}
\label{tab:adjResults}
\end{table*}

%

We evaluate \ourmethod~and \ourbaseline~using 10-fold cross validation on DREAM4 networks. \ourbaseline~uses the same rules as our approach but computes $\textsc{StandardAdj}$ as ground $\textsc{Dep}(A, B)$ atoms that never appear in groundings of $\textsc{CondIndep}(A, B, S)$, based on definition.

To evaluate the additional benefit of joint rules, we compare sub-models of \ourmethod~and \ourbaseline~run with causal orientation rules only, denoted \pslpc~and \baselinepc~respectively.
Table \ref{tab:results} shows average $F_{1}$ scores of all model variants for the regulatory network prediction task on DREAM4.
\paragraph{Noisy Extractions Maintain Performance} 
First, we see comparable performance between \ourmethod~and \ourbaseline, answering our first experimental question on how closely noisy extractions approximate adjacencies. 
In table \ref{tab:results}, there is no statistically significant difference between the $F_{1}$ scores of \ourmethod~and \ourbaseline.
The comparable performance between \ourmethod~and \ourbaseline suggests that the noisy extractions can substitute observational data computations without significantly degrading performance.

\paragraph{Joint Rules Overcome Noise}
Our investigation into model variants sheds light on the second experimental question around how logical rules overcome the noise from extractions. 
When comparing PC-only variants of each method, \baselinepc~gains over \pslpc, suggesting that sophisticated joint rules are needed to mitigate the noise from KB extractions.
The consistency across predictions encouraged by the joint rules bolsters the extraction-based adjacency signal.

\paragraph{Extractions Yield Higher Precision, Lower Recall}
To further investigate the extraction evidence mined from STRING, we compare both \textsc{StandardAdj} and \textsc{TextAdj} against gold-standard adjacencies, which we obtain from undirected regulatory links. Table \ref{tab:adjResults} shows the average precision and recall of each adjacency evidence type across the DREAM4 subnetworks. Interestingly, \textsc{TextAdj} achieves higher precision than its statistical counterpart. However, \textsc{StandardAdj} gains over \textsc{TextAdj} in recall.
The result further substantiates the benefit of joint modeling in recovering additional orientations under low-recall inputs.
Nonetheless, the comparison points to the need for a deeper understanding of the role KBs play in causal reasoning.

\subsection{Experiment Details}
\label{subsec:setup}
To obtain marginal and conditional (in)dependence tests, we use linear and partial correlations with Fisher's Z transformation. We condition on all sets up to size two.
We set rule weights for both PSL models to 5.0 except for rule C2 which is set to 10.0, since it encodes a strong acyclicity constraint.
Both models use an $\alpha$ threshold on the $p$-value to categorize independence tests as \condassocpred~or \condindeppred. We select $\alpha$ with 10-fold cross validation.
We hold out each subnetwork in turn and use the best average $F_{1}$ score across the other subnetworks to pick $\alpha \in \{0.1, 0.05\}$ raised to powers ${\{1,2,3,4,5\}}$. 
\ourbaseline~selects two different $\alpha$ values for binning independence tests and computing adjacencies, and \ourmethod~requires a single $\alpha$ for tests only. 
We also select rounding thresholds for both PSL models within the same cross-validation framework.
Since $\alpha$ is typically small, we rescale truth values $p$ for \condindeppred~by $\sqrt[3]{p}$ to reduce right-skewness of values.
We rescale all STRING affinity scores to be between 0 and 1.
\section{Related Work}
\label{sec:related}

Our work extends constraint-based methods to \cd, most notably the PC algorithm \cite{spirtes1991algorithm}, which first infers adjacencies and maximally orients them using deterministic rules based on conditional independence.
PC only supports external evidence in the form of fixed edges or non-edges.
Our work is motivated by recent approaches that cast \cd~as a SAT instance over conditional independence statements \cite{hyttinen2014constraint,magliacane2016ancestral,hyttinen2013discovering}. SAT-based approaches are based on logical representations that more readily admit additional constraints and relations from domain knowledge. However, so far, logical \cd~methods use external evidence to identify probable edges.

In a separate vein, prior work has extended text-mining to identify regulatory networks and genetic interactions only from scientific literature \cite{rodriguez2007automatic,song2009text,poon2014literome}. In contrast, our goal is to propose techniques that leverage both statistical test signals and text evidence. 
The work most similar to ours combines gene expression data with evidence mined from knowledge bases to infer gene regulatory networks \cite{chouvardas2016inferring}. However, the regulatory network inference orients edges using hard-coded knowledge of transcription factors instead of reasoning about causality. In our approach, we propose a principled causal discovery formulation as the basis of incorporating KB evidence.
\section{Discussion and Future Work}
\label{sec:discussion}

In this work, we present an initial approach for reasoning with noisy extraction-based evidence directly in a logical \cd~system. We benefit from a flexible logical formulation that supports replacing conventional adjacencies computed from observational data with cheaply obtained extractions.
Our evaluation suggests that the noisy KB-based proxy signal achieves comparable performance to conventional methods. The promising result points to future research in exploiting KBs for causal reasoning, greatly mitigating the need for costly observational data.
We see many directions of future work, including better extraction strategies for mining scientific literature and finding text-based proxies for additional statistical test signals. KBs could provide ontological constraints or semantic information useful for causal reasoning. We additionally plan to study knowledge-based constraints for causal discovery.
\bibliographystyle{plainnat}
\bibliography{causal}

\end{document}